\documentclass[letterpaper, 10pt, conference]{style/ieeeconf} 
\makeatletter
\let\NAT@parse\undefined
\makeatother
\usepackage{dblfloatfix}
\usepackage[numbers, sectionbib, sort]{natbib}
\usepackage{bm}
\usepackage{gensymb}
\usepackage{xcolor}
\usepackage{graphicx}
\usepackage{amsmath}
\usepackage{amssymb}
\usepackage{subcaption}
\usepackage{amsfonts}
\usepackage{siunitx}
\usepackage{booktabs}
\usepackage{makecell}
\usepackage{multirow}
\usepackage{upgreek}
\usepackage[font=small]{caption}
\usepackage[export]{adjustbox}
\usepackage{tikz}
\usepackage{tabularx}
\usepackage{sidecap} \sidecaptionvpos{figure}{c}
\usepackage[hidelinks]{hyperref}
\usepackage[nameinlink, capitalize]{cleveref}
\usepackage[printonlyused,withpage,nolist,nohyperlinks]{acronym}

\Crefname{section}{Sec.}{Sec.}
\Crefname{equation}{Eq.}{Eq.}

\newcommand{\etal}{\textit{et al}.}

\newcommand{\eg}{\textit{e}.\textit{g}.}

\IEEEoverridecommandlockouts

\title{\Large \bf Safe Leaf Manipulation for Accurate Shape and Pose Estimation of Occluded Fruits}

\author{Shaoxiong Yao$^\star$ \and Sicong Pan$^\star$ \and Maren Bennewitz \and Kris Hauser%
\thanks{$^\star$These authors contributed equally to this work.}
\thanks{S. Yao and K. Hauser are with the University of Illinois at Urbana-Champaign, IL, USA.
S. Pan and M. Bennewitz are with the Humanoid Robots Lab, University of Bonn, Germany.
M. Bennewitz and S. Pan are additionally with the Lamarr Institute for Machine
Learning and Artificial Intelligence and the Center for Robotics, Bonn, Germany.
This work has partially been funded by the USDA/NIFA Grant \#2020-67021-32799, by the DFG grant 459376902 – AID4Crops, under Germany’s Excellence Strategy, EXC-2070 – 390732324 – PhenoRob, and by the BMBF within the Robotics Institute Germany, grant No. 16ME0999.}\
}

\begin{document}

\maketitle
\thispagestyle{empty}
\pagestyle{empty}
\begin{abstract}

Fruit monitoring plays an important role in crop management, and rising global fruit consumption combined with labor shortages necessitates automated monitoring with robots.
However, occlusions from plant foliage often hinder accurate shape and pose estimation. 
Therefore, we propose an active fruit shape and pose estimation method that physically manipulates occluding leaves to reveal hidden fruits.  
This paper introduces a framework that plans robot actions to maximize visibility and minimize leaf damage. We developed a novel scene-consistent shape completion technique to improve fruit estimation under heavy occlusion and utilize a perception-driven deformation graph model to predict leaf deformation during planning.
Experiments on artificial and real sweet pepper plants demonstrate that our method enables robots to safely move leaves aside, exposing fruits for accurate shape and pose estimation, outperforming baseline methods.
Project page: \url{https://shaoxiongyao.github.io/lmap-ssc/}.
\end{abstract}

\section{Introduction} \label{S:introduction}

Fruits are experiencing a surge in global consumption driven by a growing population~\citep{beed2021fruit}, but current production levels are insufficient to meet future needs and labor shortages hinder the expansion of production efforts. Automation using agricultural robots is a promising approach for enhancing production and satisfying the escalating demand, as robots can take over labor-intensive tasks, such as selective harvesting~\citep{lenz2024hortibot} and crop monitoring~\citep{marangoz2022case}. Crop monitoring involves estimating count, growth, pose, appearance, and disease occurrence, which helps with production forecasting or pesticide and fertilizer application. 
Automated monitoring is difficult due to occlusions, which are primarily caused by plant foliage.
Recent advances in artificial intelligence are able to mitigate this issue~\citep{storm2024research} by estimating the shapes and poses of occluded fruits using shape completion techniques~\cite{marangoz2022case,pan2023panoptic,magistri2024icra}. 
However, these approaches still suffer from inaccuracies in the presence of heavy occlusion.

\begin{figure}[!t]
\centering
\includegraphics[width=0.95\columnwidth]{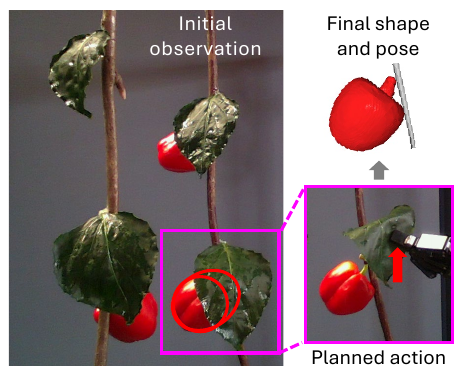}
\caption{Example of our system in a setting with sweet peppers.
Left: The zoomed-in area in the initial observation showcases the occlusion of the targeted pepper. 
The red profiles represent different possible shapes and poses of the fruit consistent with the visible portion. 
Right: Our method selects an appropriate manipulation action to reveal the entire fruit while preserving the leaf's integrity.
This enhanced visibility enables accurate shape and pose estimation.
}
\label{fig:StarFigure}
\vspace{-0.5cm}
\end{figure}

We propose to use physical manipulation to move aside occluding leaves to directly observe the hidden fruit.  
Our active fruit shape and pose estimation system integrates novel visibility and deformation simulation methods to predict safe and effective manipulation actions. 
It consists of three core technical contributions. 
First, we develop a Scene-consistent Shape Completion~(SSC) method that uses semantic-aware and free space prior to improve the fruit shape and pose estimates beyond existing methods~\cite{marangoz2022case,pan2023panoptic,magistri2024icra} in highly occluded settings.
Second, to decide on an appropriate manipulation action, we found that heuristic approaches are prone to damage the plant by tearing the leaf or breaking it at the stem. 
Instead, our Leaf Manipulation Action Planning~(LMAP) approach uses a perception-driven, energy-based deformation graph model~\cite{sumner2007siggraph} to predict plant state and energy for candidate actions. 
Third, we integrate the shape completion and deformation prediction together in an integrated pipeline. 
Together, these innovations enable our robot to generate actions that move the leaf aside to observe the fruit while simultaneously reducing the risk of damage to the plant. 
Fig.~\ref{fig:StarFigure} illustrates an example of our pipeline safely removing the occlusion of a sweet pepper, resulting in accurate shape and pose estimation.

We validate our approach in experiments on an artificial sweet pepper plant setting with varying leaf sizes and fruit shapes, poses, and colors, demonstrating the practical applicability of our approach and superiority compared to baselines.
Our method is consistently able to select occlusion-removal manipulations that reveal the entire visible fruit surfaces with a low risk of plant damage, enabling better shape and pose estimation.
Experiments with real leaves and peppers further highlight the potential for realistic agricultural applications, such as fruit monitoring in glasshouses.

\section{Related Work} \label{S:related_work} 

\subsection{Autonomous Agriculture}

There is a growing research trend towards enhancing autonomy in precision agriculture.
Approaches include designing specialized hardware~\cite{uppalapati2020berry}, developing autonomous systems~\cite{smitt2021pathobot}, and advancing perception algorithms~\cite{smitt2023pag}.
Among these, the most relevant literature for this work pertains to active perception algorithms that reduce occlusions by moving the camera to gather informative observations based on viewpoint planning methods~\cite{zeng2020view}.
Zaenker \etal~\cite{zaenker2021combining,zaenker2021viewpoint,zaenker2023graph} propose viewpoint planning methods for maximizing the coverage of the region of interests, \eg, fruit areas. 
Burusa \etal~\cite{burusa2024attention,burusa2023efficient} utilize attention-driven and semantic-aware view planning for reconstruction and detection of relevant plant parts. 
However, these will not handle heavy occlusions, as there are no viewpoints that provide sufficient visibility, and the robot has to physically move the leaf aside.

\subsection{Interactive Perception}

A robot can explore an environment and gain informative observations by interacting with objects~\cite{bohg2017interactive}. 
In cluttered environments like shelves, a robot can reveal occluded objects by moving front objects aside~\citep{li2016iros,danielczuk2019mechanical,huang2021mechanical,dengler2023iros}. 
Model-free reinforcement learning approaches can generate diverse exploration actions but require simulation to generate sufficient training data~\citep{cheng2018corl}. 
Our method is closer to model-based approaches that predict action effects and maximize uncertainty reduction~\citep{huang2021mechanical}.
Compared to shelf environments, plants introduce additional complications of deforming geometry and uncertain connectivity between parts. 

The model-based approach requires a simulator of the plant; however, existing work is limited to predicting behavior with a perfect skeleton~\cite{deng2024gazebo} or simulating contact forces~\cite{yao2023icra,yao2024structured}.
In the context of manipulation action planning on plants, existing methods avoid constructing a precise deformation model by either employing self-supervised learning~\citep{zhang2023push} or pushing on branches~\citep{gursoy2024occlusion}. 
Zhang and Gupta~\citep{zhang2023push} suggest a data-driven push planning method for plant space revealing. 
However, their method requires extensive time for collecting training data for a particular scenario (over 30 hours for vines and over 18 hours for Dracaena), which makes it difficult to generalize to unseen scenarios and plants.
Additionally, their approach focuses solely on space revealing rather than addressing the task of fruit monitoring.
Gursoy \etal~\citep{gursoy2024occlusion} propose a branch push planning method for occlusion handling in fruit detection.
However, their method considers just ellipses for fruit shape estimation.
Additionally, in most cases, fruit occlusion is primarily caused by leaves, which cannot be easily resolved by simply moving branches.
For our setting, we use an energy-based deformation graph model~\cite{sumner2007siggraph} that is constructed from the semantic segmentation of the observed point cloud. 

\subsection{Shape Completion}

A typical shape completion model generates a completed shape of the object based on partial observation~\cite{yu2021pointr}. 
Such techniques are widely used in viewpoint planning to reduce occlusions during online reconstruction of both single objects~\cite{pan2022scvp,dhami2023pred,liu2024nbv} and plants~\cite{wu2019plant,menon2023nbv}.
For fruit monitoring, the shape completion methods usually take the pose of the fruit into account as well.
Marangoz \etal~\cite{marangoz2022case} suggest using superellipsoids to fit the posed fruit shape based on current observations.
Magistri \etal~\cite{magistri2022contrastive,magistri2024icra,magistri2024ral} propose using deep learning-based methods for sweet pepper and strawberry shape completion.
Pan \etal~\cite{pan2023panoptic} consider surface consistency of DeepSDF~\cite{park2019deepsdf} and occlusion-aware differentiable rendering for fruit shape and pose estimation. 
However, these methods overlook the semantic information of the scene, which is a key aspect we address in this work.

\section{Method} \label{S:our_approach}

\begin{figure}[!t]
\centering
\includegraphics[width=1.0\columnwidth]{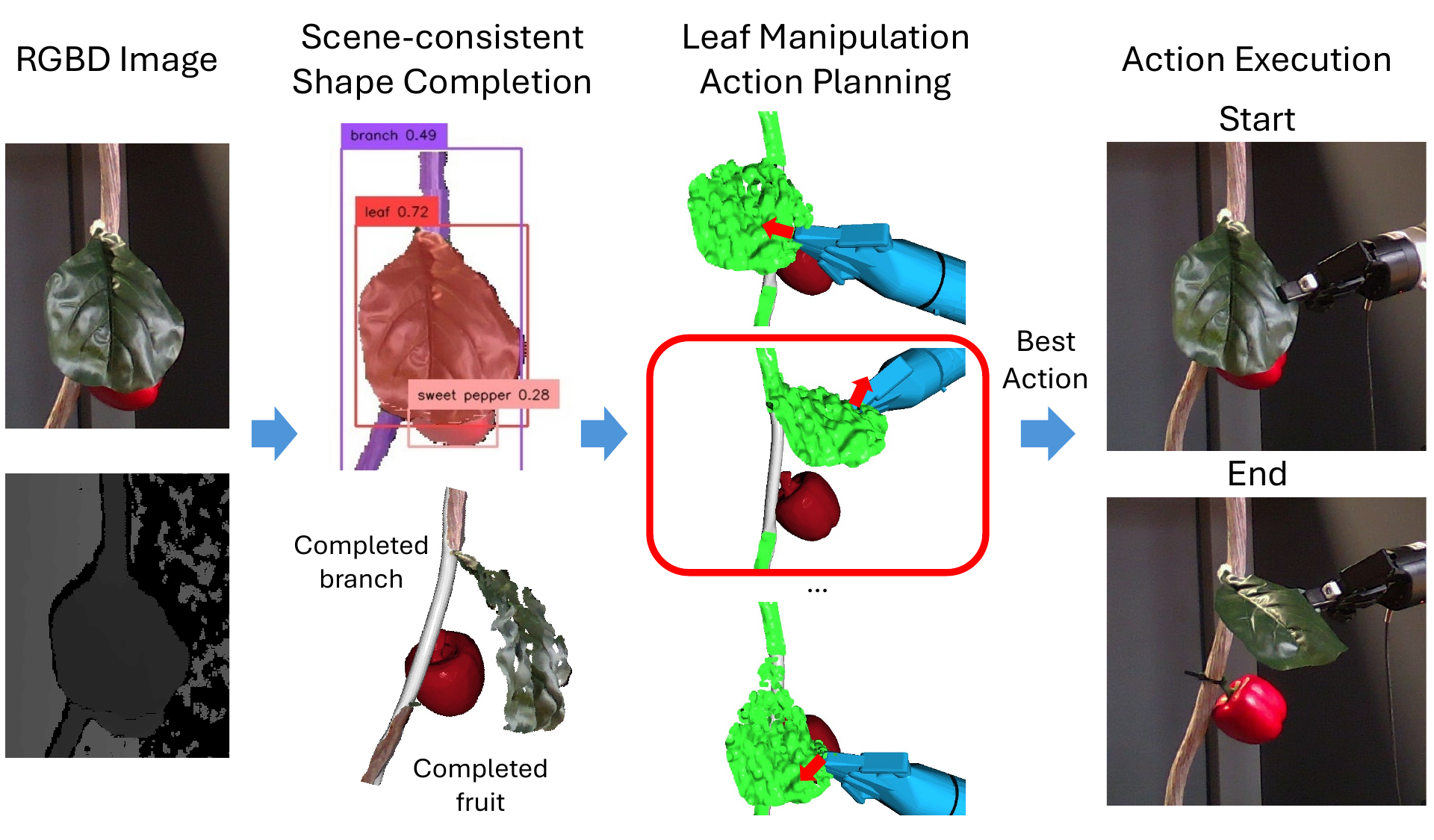}
\caption{
Overview of the proposed active fruit shape and pose estimation system for plant-safe occlusion removal. 
Given an \mbox{RGB-D} image, a semantic representation informs our scene-consistent shape completion module to build an initial estimate of the fruit shape and pose. 
The deformation prediction is performed on the same semantic representation for acquiring the deformed states of the leaf, while the action planner predicts fruit visibility and damage risk for several candidate actions, and the best manipulation is executed.
After execution, the fruit's shape and pose are re-estimated using the updated observation information.
}
\label{fig:OverviewFigure}
\vspace{-0.5cm}
\end{figure}

An overview of our active fruit shape and pose estimation system is illustrated in Fig.~\ref{fig:OverviewFigure}.
The system is comprised of three components: scene-consistent shape completion, deformation prediction, and leaf manipulation action planning.
All components are perception-based and have no prior model of the plant, although the shape completion module is assumed to have a 3D template that has been trained on scanned fruits of the same species and similar growth stages.
As the input to the system, we assume only an RGB-D image taken from a camera with a known pose relative to the robot arm base.
The shape completion module then segments and estimates the 3D shape of the branch and fruit, using scene consistency as prior knowledge.
Afterwards, the deformation model is instantiated on the completed scene, predicting how it will deform under candidate grasp-and-pull actions.
Based on these predictions, our action planning system selects the best action that maximizes fruit visibility while reducing the risk of damaging the plant.
The planned grasp and pull action is then executed on the real robot.
Following the execution, we re-estimate the fruit's shape and pose based on the updated observational data.

\subsection{Scene-Consistent Shape Completion (SSC)}

The key to our scene-consistent shape completion (SSC) is incorporating free-space and peduncle information.
We assume that the peduncle is attached to the branch that may also be occluded by leaves.
To address this issue, we first segment the scene from an RGB-D input and perform branch completion before integrating the data into our fruit shape completion method.

\subsubsection{Semantic Segmentation}

We segment the RGB image using a pre-trained open-vocabulary segmentation model, GroundedSAM~\cite{ren2024grounded}.  
We exclude RGB pixels with a large depth value ($> 1.5$\,m in our setup) before feeding them into the model.
We use the text prompt \verb|[branch, leaf, sweet pepper]| to segment the corresponding parts of the plant into instances.
The segmented pixels are de-projected into 3D space using the camera intrinsic and depth image to obtain the semantic point clouds, denoted as $P_{\text{branch}}$, $P_{\text{leaf}}$, and $P_{\text{fruit}}$.

\subsubsection{Branch Completion}

To ensure a continuous shape for the occluded part of the branch, we deform a cylinder triangle mesh to $P_{\text{branch}}$ using as-rigid-as-possible (ARAP) mesh deformation~\cite{sorkine2007arap}. 
First, we run principal components analysis~(PCA) on $P_{\text{branch}}$ to find the principal axis of the branch. 
Next, a triangle mesh for the cylinder is initialized by connecting the two end points along the principal axis.
The cylinder's radius is determined by averaging the radial distances measured across a series of planes along the principal axis.
We then deform this initial cylinder triangle mesh to match $P_{\text{branch}}$. 
During each deformation iteration, we match point cloud points only to visible mesh vertices within a distance threshold to prevent fitting onto invisible vertices.
These corresponding points are then used as handles in ARAP mesh deformation. 
The optimization process terminates when the ARAP energy no longer decreases. 
The completed mesh is denoted as $M_{\text{branch}}$.

\subsubsection{Fruit Shape Completion}

We adopt the surface representation of the DeepSDF template of Pan \etal~\cite{pan2023panoptic}, pretrained on scanned 3D models of sweet peppers.
DeepSDF~\cite{park2019deepsdf} takes as input a query $\mathrm{x} \in \mathbb{R}^3$ and a latent shape code $\mathrm{z} \in \mathbb{R}^{C}$ ($C=32$ as in~\cite{pan2023panoptic}), and predicts the SDF value $v \in \mathbb{R}^3$ at $\mathrm{x}$ through a decoder network $D_{\theta}(\mathrm{x},\mathrm{z})$.

We jointly estimate the fruit's latent shape code $\mathrm{z}$ and 7 DOF pose transformation $\mathrm{T_{w\rightarrow o}} \in Sim(3)$. $\mathrm{T_{w\rightarrow o}}$ maps from the robot world coordinate system to the fruit’s canonical coordinate system, consisting of scale $\mathrm{s} \in \mathbb{R}$, translation $\mathrm{t} \in \mathbb{R}^3$ and rotation $\mathrm{R} \in \mathbb{R}^{3\times3}$.
We optimize the standard surface loss~\cite{pan2023panoptic} but introduce novel scene consistency objectives as follows.

\textbf{Surface loss}: The standard surface loss $\mathcal{L}_{\text{sur}}$  objective attracts the iso-surface of the SDF predicted by $D_{\theta}$ toward points on the observed fruit surfaces $P_{\text{fruit}}$:
\begin{equation}
\mathcal{L}_{\text{sur}} = \frac{1}{\left|P_{\text{fruit}}\right|} \sum_{{p}^{w} \in P_{\text{fruit}}} D_\theta^2(T_{w\rightarrow o} {p}^{w}, \mathrm{z}).
\label{eq:loss_sur}
\end{equation}

\textbf{Negative-point loss}: The second objective is to avoid surface collision with negative points $P_{\text{neg}}$ consisting of $P_{\text{leaf}}$, $P_{\text{branch}}$, sampled points from $M_{\text{branch}}$, and the free space points $P_{\text{free}}$. 
$P_{\text{free}}$ is computed from the voxelized scene using OctoMap~\citep{hornung2013auro}, with a default resolution of 3\,mm.
The loss $\mathcal{L}_{\text{neg}}$ pushes the iso-surface of the SDF predicted by $D_{\theta}$ such that negative points remain in the fruit exterior:
\begin{equation}
\mathcal{L}_{\text{neg}} = \frac{1}{\left|P_{\text{neg}}\right|} \sum_{{p}^{w} \in P_{\text{neg}}} \max(0, -D_\theta(T_{w\rightarrow o} {p}^{w}, \mathrm{z})).
\label{eq:loss_neg}
\end{equation}

\textbf{Peduncle loss}: The third objective attracts the peduncle of the fruit iso-surface to the completed branch.
As in the pretrained template~\cite{pan2023panoptic}, the peduncle always aligns with the vertical axis in the fruit’s canonical coordinate system.
We obtain a set number of points (default is 10) with the highest $z$ values to represent the peduncle points $P_{\text{ped}}$. 
The nearest corresponding points $P_{\text{cor}}$ are computed by searching among the sampled points from $M_{\text{branch}}$.
We keep $P_{\text{ped}}$ close to $P_{\text{cor}}$, which minimizes the peduncle loss $\mathcal{L}_{\text{ped}}$ given by:
\begin{equation}
\mathcal{L}_{\text{ped}} = \frac{1}{\left|P_{\text{ped}}\right|} \sum_{\left({p}^{o},{p}^{w}\right) \in \left(P_{\text{ped}},P_{\text{cor}}\right)} \left\| T_{w \rightarrow o}^{-1} {p}^{o} - {p}^{w} \right\|^2.
\label{eq:loss_ped}
\end{equation}

With an additional shape code regularization term $\mathcal{L}_{\text{reg}}=||\mathrm{z}||^2$, we consider our scene-consistent loss function:
\begin{equation}
\mathcal{L} = w_{\text{sur}} \mathcal{L}_{\text{sur}} + w_{\text{neg}} \mathcal{L}_{\text{neg}} + w_{\text{ped}} \mathcal{L}_{\text{ped}} + w_{\text{reg}} \mathcal{L}_{\text{reg}}.
\label{eq:loss_total}
\end{equation}
where $w_{\text{sur}},w_{\text{neg}},w_{\text{ped}},w_{\text{reg}}$ are the weights for each loss term. For the initial optimization, the weights are set to $1.0$, $2.0$, $0.1$, and $1.0$, respectively. 
Since the points sampled from the completed branch are not directly observed by the sensor and may contain noise, we assign a different weight of $0.1$ to those negative completed branch points.

We use the Adam optimizer~\citep{kingma2014adam} with PyPose~\citep{wang2023pypose} to solve this optimization problem.
The latent shape code $\mathrm{z}$ is initialized from the pretrained DeepSDF~\cite{pan2023panoptic}, $\mathrm{s}$ is initialized to $1.0$, $\mathrm{t}$ is initialized to the center of the observed fruit points $P_{\text{fruit}}$, and $\mathrm{R}$ is initialized to the identity matrix $I$.

\subsection{Perception-Based Deformation Model}

Given the completed scene, we instantiate a leaf deformation model to predict its deformed states in response to a given robot action.
An action $a = (p^g, R^g, \delta p^g)$ consists of a grasp pose $(p^g, R^g)$ and a displacement $\delta p^g$.
We use grasping over pushing for more stable contact points~\cite{zhang2023push}, improving deformation precision.
$p^g \in \mathbb{R}^3$ is the grasped point and $R^g \in SO(3)$ defines gripper orientation.
$\delta p^g \in \mathbb{R}^3$ can be calculated using a pull direction and a move distance.

To simulate the leaf deformation using~\citep{sumner2007siggraph}, three types of points need to be defined: 
(1) Fixed points $P_{fix}$ served as the base. 
We assume the branch is much stiffer than the leaf and remains fixed.
(2) Points $P_{def}$ where deformation needs to be predicted. 
These are observed leaf points.
(3) Points $P_{gri}$ with known deformation. 
Our action $a$ provides known deformation of the grasping point.
As the gripper's geometry provides a more informative gripping area instead of point, we simulate the area by a 5×5 grid of points $P_{gri}$ within a 2.5\,cm square calculated using robot forward kinematics.

We define the embedded deformation graph $G = (P, \mathcal{E})$ using these types of points, where $P = P_{fix} \cup P_{def} \cup P_{gri}$ and edges~$\mathcal{E}$ are formed by connecting points that are within a distance threshold (default is 2\,cm).
As in~\cite{sorkine2007arap}, for rigid deformation $P\rightarrow P'$ there exists a rotation matrix $R'_i$ for each $p_i$ and deformed $p'_i$ with its all edges $(i,j)\in \mathcal{E}$:
\begin{equation}
    \begin{aligned}
    p'_i - p'_j = R'_i(p_i - p_j),\ \forall (i,j) \in \mathcal{E}.
    \end{aligned}
    \label{eq:rigid_rotation}
\end{equation}
The ARAP energy is defined by the non-rigid deformation that can be approximated in a weighted least squares sense:
\begin{equation}
    \begin{aligned}
    E(P, P') = \sum_{(i, j) \in \mathcal{E}} w_{ij} \cdot \| p'_i - p'_j - R'_i(p_i - p_j)\|^2,
    \end{aligned}
    \label{eq:EdgEnergy}
\end{equation}
where $w_{ij}$ represents the connection weight.
The center line of the leaf is harder due to the veins, which provide structural support.
We extract the center line using PCA and points $P_{vei} \subset P_{def}$ belonging to the veins are defined as those with a distance smaller than a threshold (default is 1.2\,cm).
If an edge has two points belonging to $P_{vei}$, set $w_{ij}=3$; if it has one point, set $w_{ij}=2$; if it has no points, set $w_{ij}=1$.
The predicted deformation $\hat{P}'$ can be optimized by minimizing:
\begin{equation}
    \begin{aligned}
        \hat{P}' & = \arg\min_{P'} E(P, P'), \\
        \text{s.t.} \quad p'_i & = p_i + \delta p_i, \ \forall p_i\in P_{gri},\\ 
        p_i' & = p_i, \quad \quad \quad \forall p_i \in P_{fix}.
    \end{aligned}
\end{equation}

To facilitate the efficiency of deformation simulation, sparse representations are desired.
We voxelize the related segments (leaf and branch) from the completed scene with a default resolution of 5\,mm.
We use the efficient global-local optimization algorithm~\cite{sorkine2007arap}.
The predicted minimized energy $\hat{E}$ also serves as a metric for assessing action safety, as actions that excessively stretch the leaf are associated with high ARAP energy values.
In contrast, safer actions that gently pull the leaf aside result in lower ARAP energy.
We finally apply radial basis function interpolation~\citep{newcombe2015dynamicfusion} of our sparse deformation $\hat{P}'$ to a dense deformation.

\subsection{Leaf Manipulation Action Planning (LMAP)}

The goal of our leaf manipulation action planning~(LMAP) is to reduce the risk of damaging the plant by energy prediction $\hat{E}$ and maximize predicted fruit visibility based on deformation states $\hat{P}'$, completed fruit mesh $M_{\text{fruit}}$ extracted from estimated $D_{\theta}$, and the robot mesh.

\subsubsection{Action Generation and Collision Checking}

Our LMAP method first samples action candidates on the edges of the leaf to avoid twisting the leaf. 
First, we project the $P_{\text{leaf}}$ from 3D to 2D using PCA. 
Then, we fit a convex hull on the 2D points where we can meaningfully define the leaf's boundary. 
We sample 15 grasp points on the edges of the convex hull with a small margin of 1\,cm towards the interior to enhance robustness.
We uniformly sample 14 pull directions in the robot coordinate system, including 6 axial and 8 diagonal directions. 
We additionally consider action directions that form an angle of less than $120^\circ$ between the line connecting the grasped points and the direction of the branch to avoid pulling the leaf away from it.
We sample move distances starting at 2\,cm, increasing in 2\,cm increments up to 8\,cm, or until the predicted fruit visibility reaches 95\% of its maximum (discussed in the next subsection).
In this way, we obtain a set of action proposals denoted as $\mathcal{A} = \{a_k(p^g_k, R^g_k, \delta p^g_k)\}$.

Given an action $a_k \in \mathcal{A}$, we use inverse kinematics to compute robot arm mesh $M_k$.
We check robot-fruit and leaf-fruit collisions using $D_{\theta}$, as well as robot-branch collision by converting the $M_{\text{branch}}$ into the SDF.
If any type of collision occurs, the action $a_k$ is discarded.

\subsubsection{Visibility Prediction}

We use the ray-casting of OctoMap (resolution 3\,mm) from the camera viewpoint to count the number of fruit surfaces hit.
The initial hits $H_{init}$ is computed by inserting $P_{\text{fruit}}$, $P_{\text{branch}}$, $P_{\text{leaf}}$ into OctoMap and counting the hits on $P_{\text{fruit}}$.
The predicted hits $H_{k}$ of action $a_k$ is computed by inserting $M_{\text{fruit}}$, deformed states $\hat{P}'_k$, robot mesh $M_k$ and counting on $M_{\text{fruit}}$.
Hence, the predicted visibility of $a_k$ is computed by $N^{vis}_k=|H_{k}\setminus H_{init}|$, indicating the increased number of visible surfaces.
The maximum visibility is computed by inserting only $M_{\text{fruit}}$.

\subsubsection{Action Selection}

The score for each action is evaluated as a weighted sum of the predicted visibility and the predicted energy, given by:
\begin{equation}
    a_k^* = \arg\max_{a_k \in \mathcal{A}} (N^{vis}_k - \lambda \cdot \hat{E}_k),
    \label{eq:ActionSelectionScore}
\end{equation}
where $\lambda$ is the weight parameter on the energy score (default is 1.0). 
$ N^{vis}_k$ and $\hat{E}_k$ are normalized over all feasible actions.
The action $a_k^*$ with the highest score is then chosen for execution, balancing visibility improvement and minimizing potential damage to the leaves.

\subsection{Action Execution and Final Estimation}

For smooth motion execution, we define a pre-grasp pose, which is set by moving a distance backward (default is 5\,cm) from the robot end-effector's grasp pose.
After executing the action, we segment the new observation and fuse the initial observation with it.
We re-estimate the shape and pose using our shape completion method with the new fused completed scene.
We assume that the final scene generally contains more informative fruit surfaces and free points.
Therefore, we reduce the peduncle weight and the negative weight of the completed branch from $0.1$ to $0.01$ in Eq.~\ref{eq:loss_total}.

\subsection{Extension to Multi-Fruit Scenarios}

Our system can be easily extended to handle multi-fruit scenarios by sequentially revealing the occlusions of several sweet peppers.
For that, branch instances are clustered using the line fitting method~\cite{lenz2024hortibot}.
Each fruit instance is assigned the occluding leaf and attached branch based on distance checking.
Subsequently, a deformation model is instantiated for each leaf along with the branch it is connected to.

\section{Experimental Results} \label{S:experimental_results}

Our experiments show that (1) fruit shape completion accuracy has a positive correlation with visibility and our scene-consistent priors improve the accuracy; (2) our planned actions achieve high visibility while reducing  plant damage risk; (3) our system can scale to scenarios with multiple real fruits and leaves. For additional experimental details, please refer to our project page or the accompanying video.

\begin{figure}[!t]
    \centering
    \scriptsize
    \begin{tabular}{ccc}
        \includegraphics[trim={1.5cm 0.2cm 0cm 0.4cm}, clip, height=2.2cm]{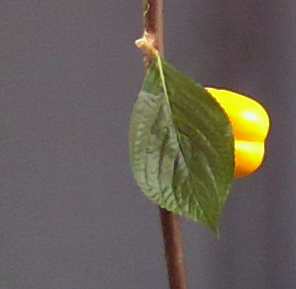} &  
        \includegraphics[trim={0cm 0.2cm 0cm 0.4cm}, clip, height=2.2cm]{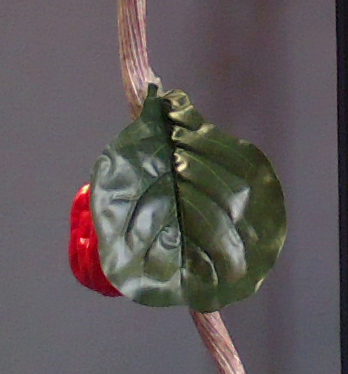} & 
        \includegraphics[height=2.2cm]{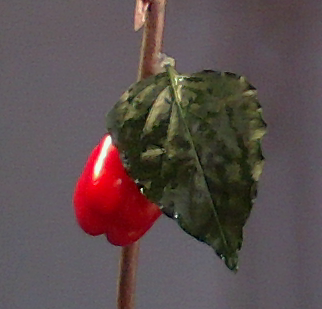}   
         \\
        (a) Small leaf & (b) Large leaf & (c) Real leaf \\
    \end{tabular}
    \caption{Representative leaf occlusion scenarios in the experiments: (a) and (b) are examples of artificial leaf and pepper configurations used in Sec.~\ref{sec:EvalShapeCompletion} and~\ref{sec:EvalActionPlanning}, respectively; (c) is an example of the real pepper leaf and real pepper configuration used in Sec.~\ref{sec:EvalRealPlant}.
    }
    \label{fig:TestCases}
    \vspace{-0.3cm}
\end{figure}

\subsection{Hardware Setup and Testing Plant Scenarios}

Our hardware setup uses a UR5 robot equipped with a Robotiq 2F-85 gripper and an calibrated eye-to-hand L515 RealSense camera.
We tested 12 artificial and 16 real plant scenarios. The artificial scenarios included two types of artificial leaves (small/large with thin/thick branches), three artificial peppers with varying shapes (circular, striped) and colors (red, yellow), and two occlusion conditions as shown in Fig.~\ref{fig:TestCases} (a) and (b). The real scenarios involved leaves and peppers with natural variations as shown in Fig.~\ref{fig:TestCases} (c). 
For quantitative evaluation, we collect ground truth shapes and poses of artificial fruits using 3D point-cloud scanning~\cite{pan2025tro} with mesh reconstruction~\cite{hu2024icra}. The fruit pose for each scenario is annotated through human labeling.

\begin{figure}[!t]
    \centering
    \scriptsize
    \setlength{\tabcolsep}{0.8pt}
    \begin{tabular}{ccccc}
    \rotatebox[origin=l]{90}{\hspace{2mm}Initial occlusion} &
    \includegraphics[clip, trim=2.0cm 0cm 2.0cm 0cm, width=0.115\textwidth]{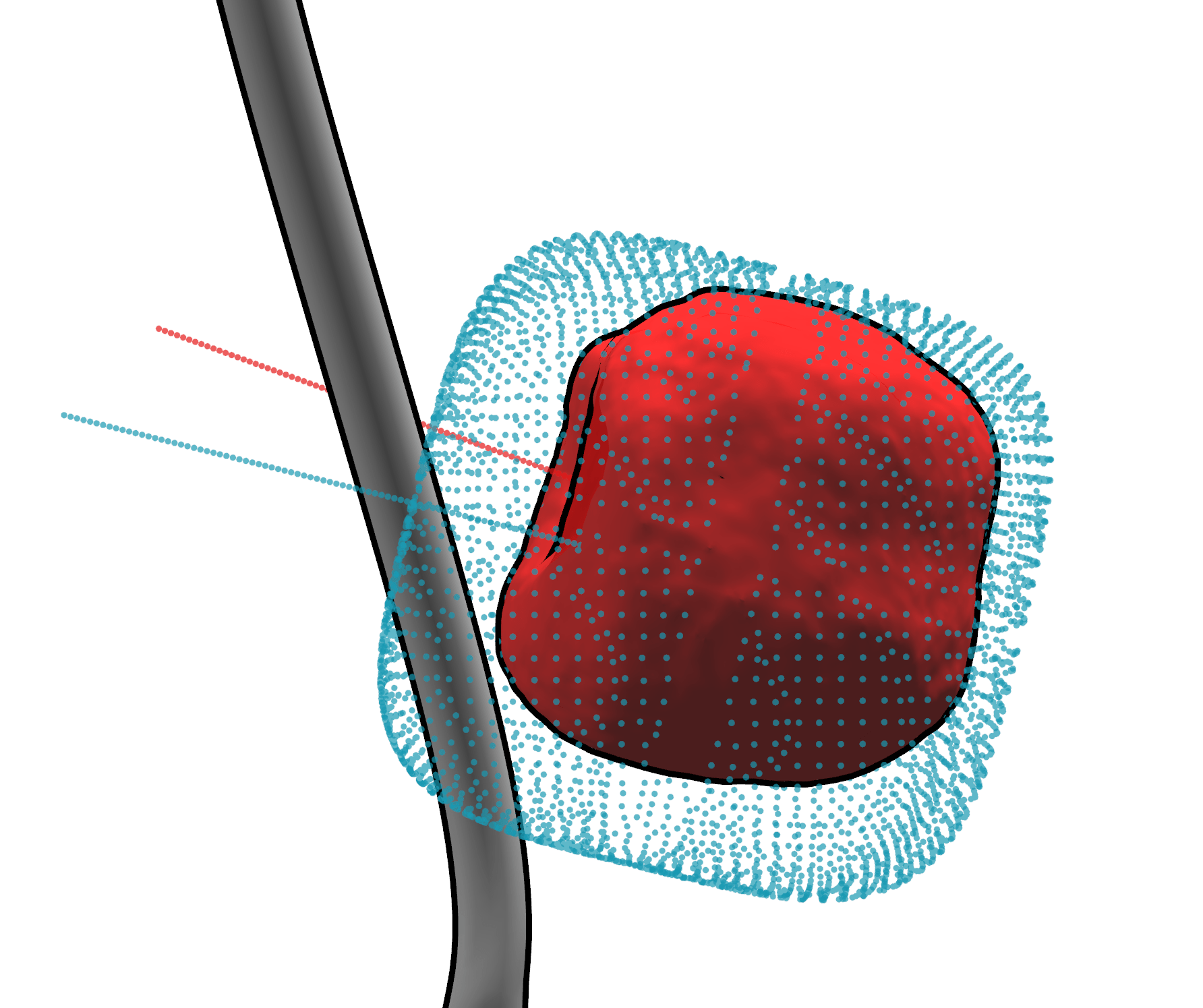} & 
    \includegraphics[clip, trim=2.0cm 0cm 2.0cm 0.5cm, width=0.115\textwidth]{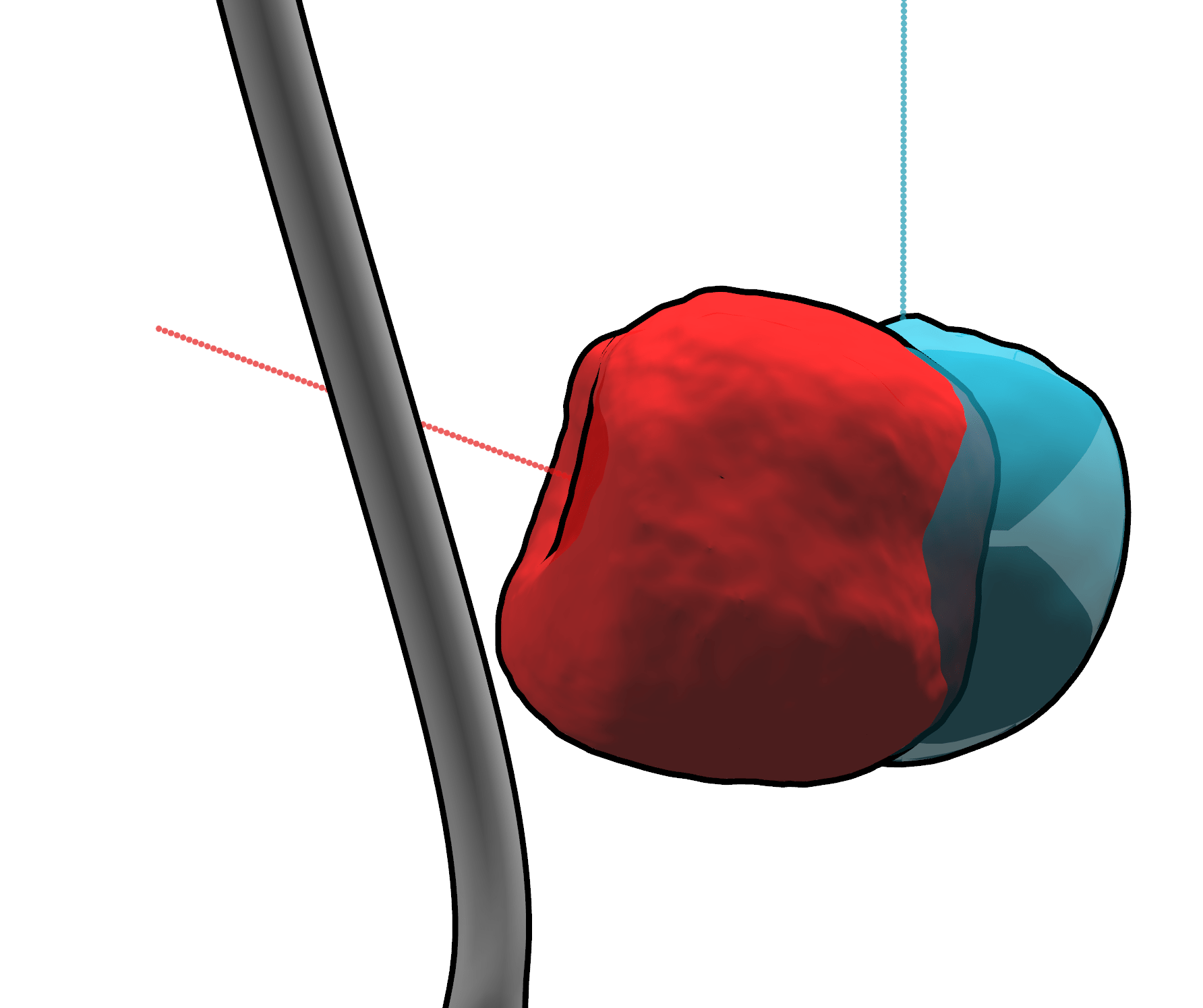} & 
    \includegraphics[clip, trim=2.0cm 0cm 2.0cm 0.5cm, width=0.115\textwidth]{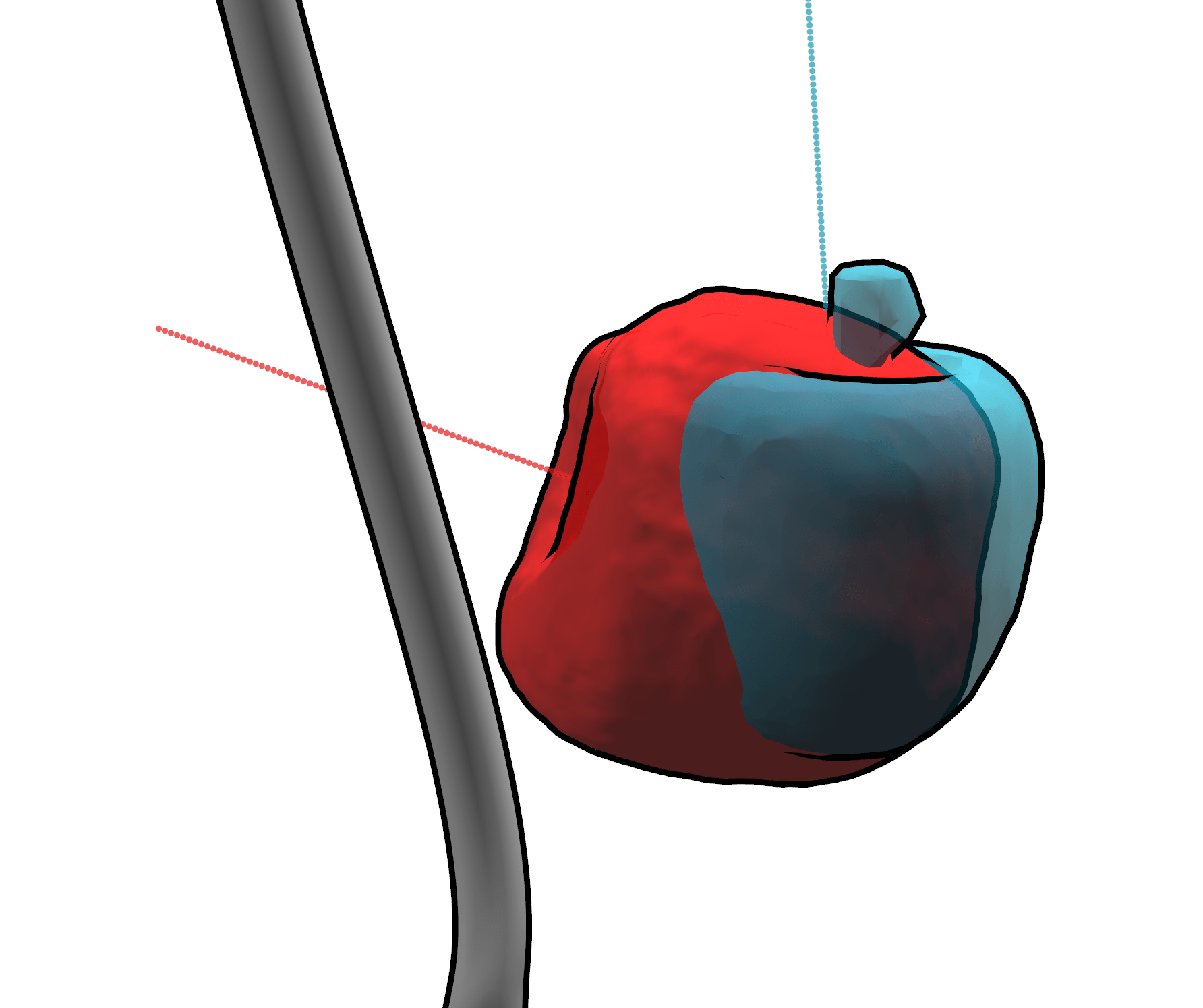} &
    \includegraphics[clip, trim=2.0cm 0cm 2.0cm 0cm, width=0.115\textwidth]{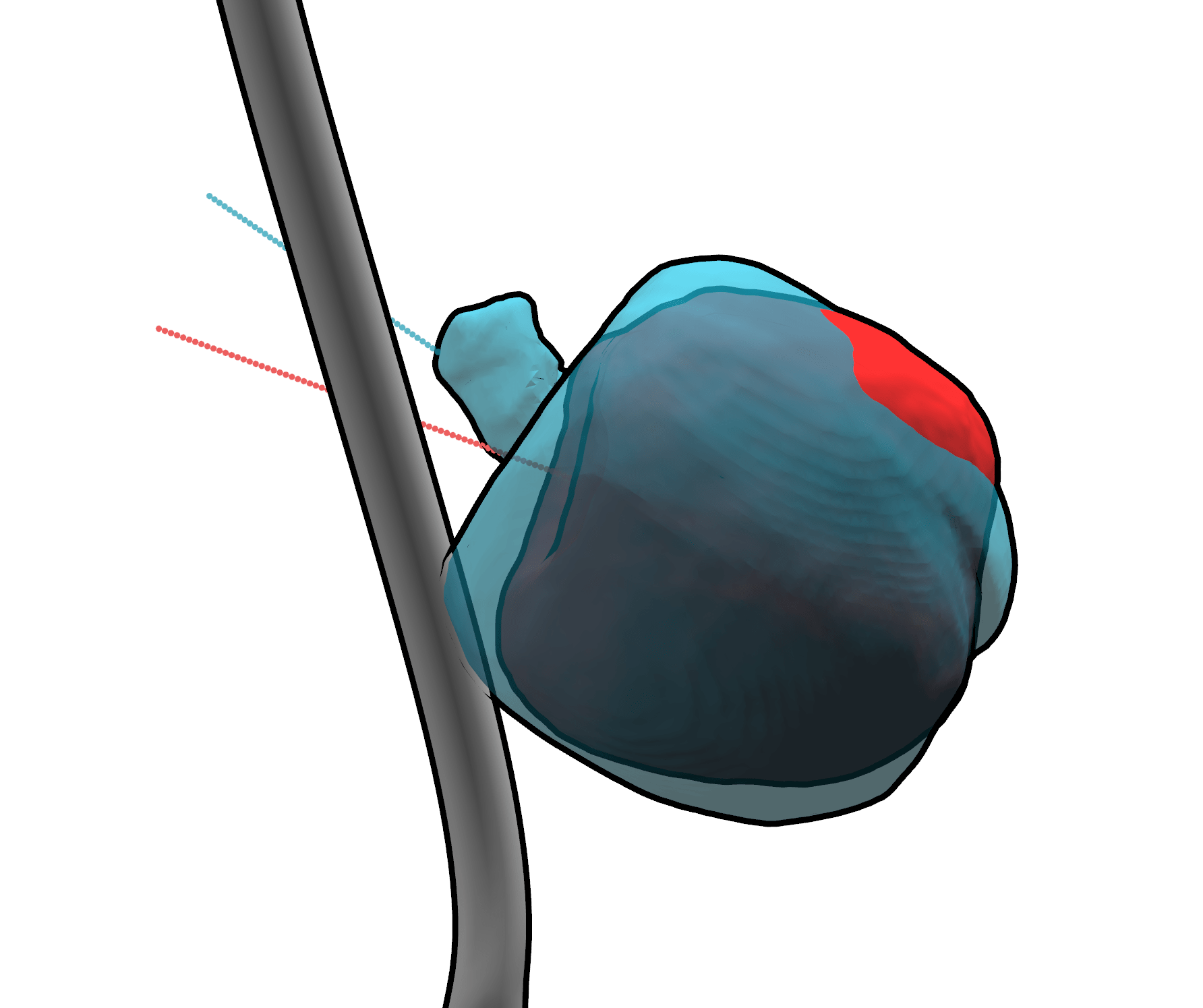} \\

    \rotatebox[origin=l]{90}{\hspace{-1mm}\shortstack{Without occlusion}} &
    \includegraphics[clip, trim=2.0cm 0cm 2.0cm 0cm, width=0.115\textwidth]{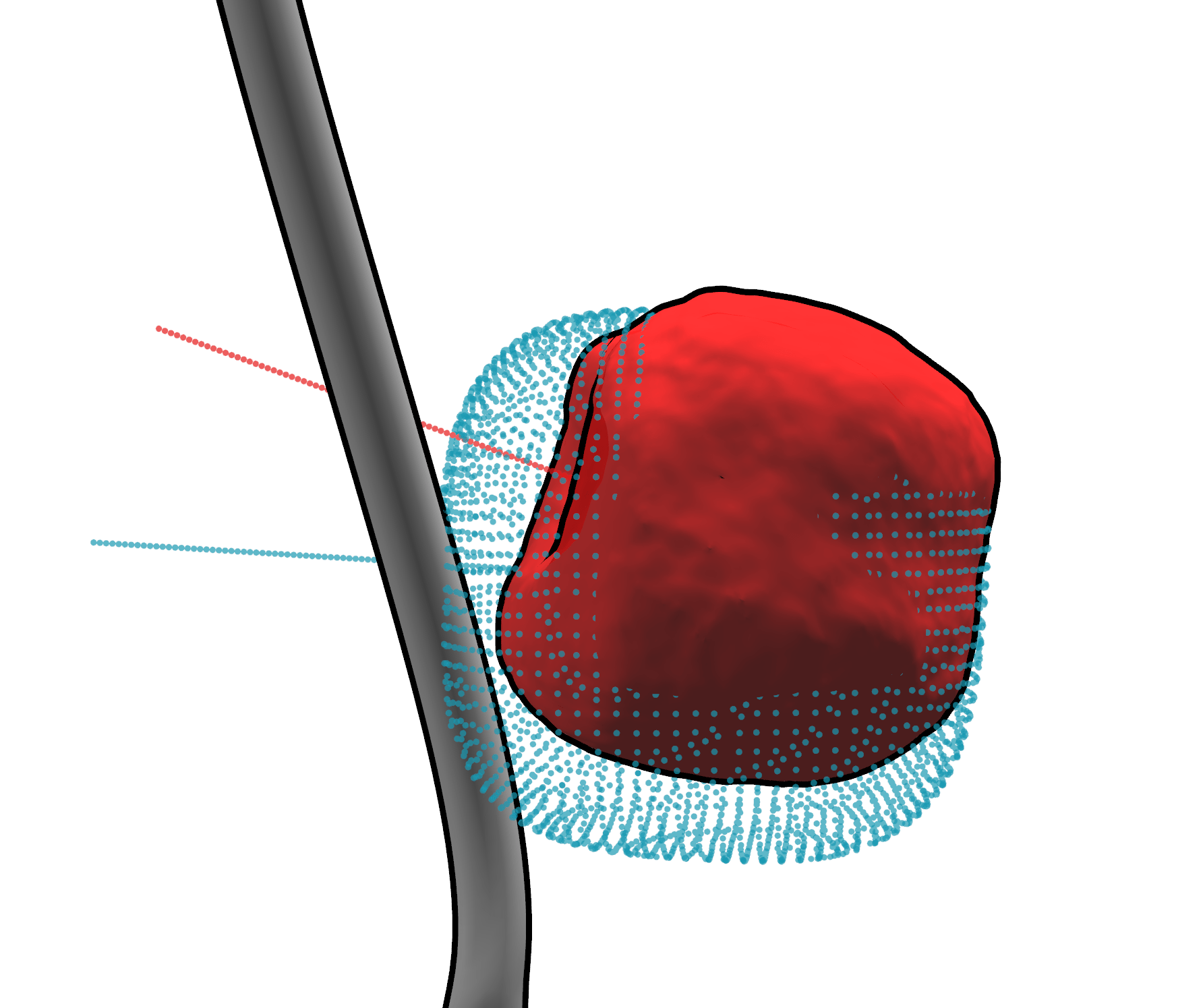} & 
    \includegraphics[clip, trim=2.0cm 0cm 2.0cm 0cm, width=0.115\textwidth]{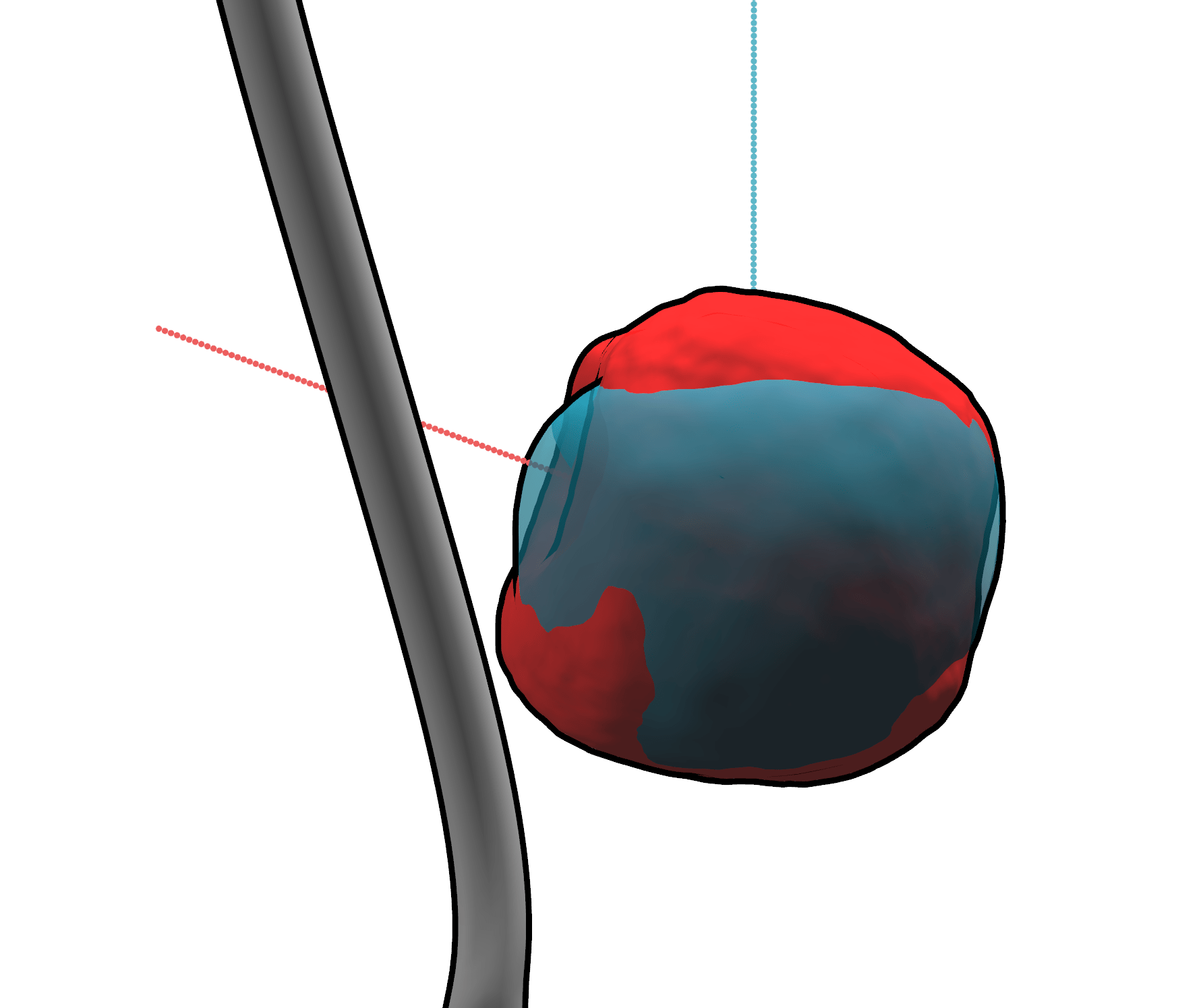} & 
    \includegraphics[clip, trim=2.0cm 0cm 2.0cm 0cm, width=0.115\textwidth]{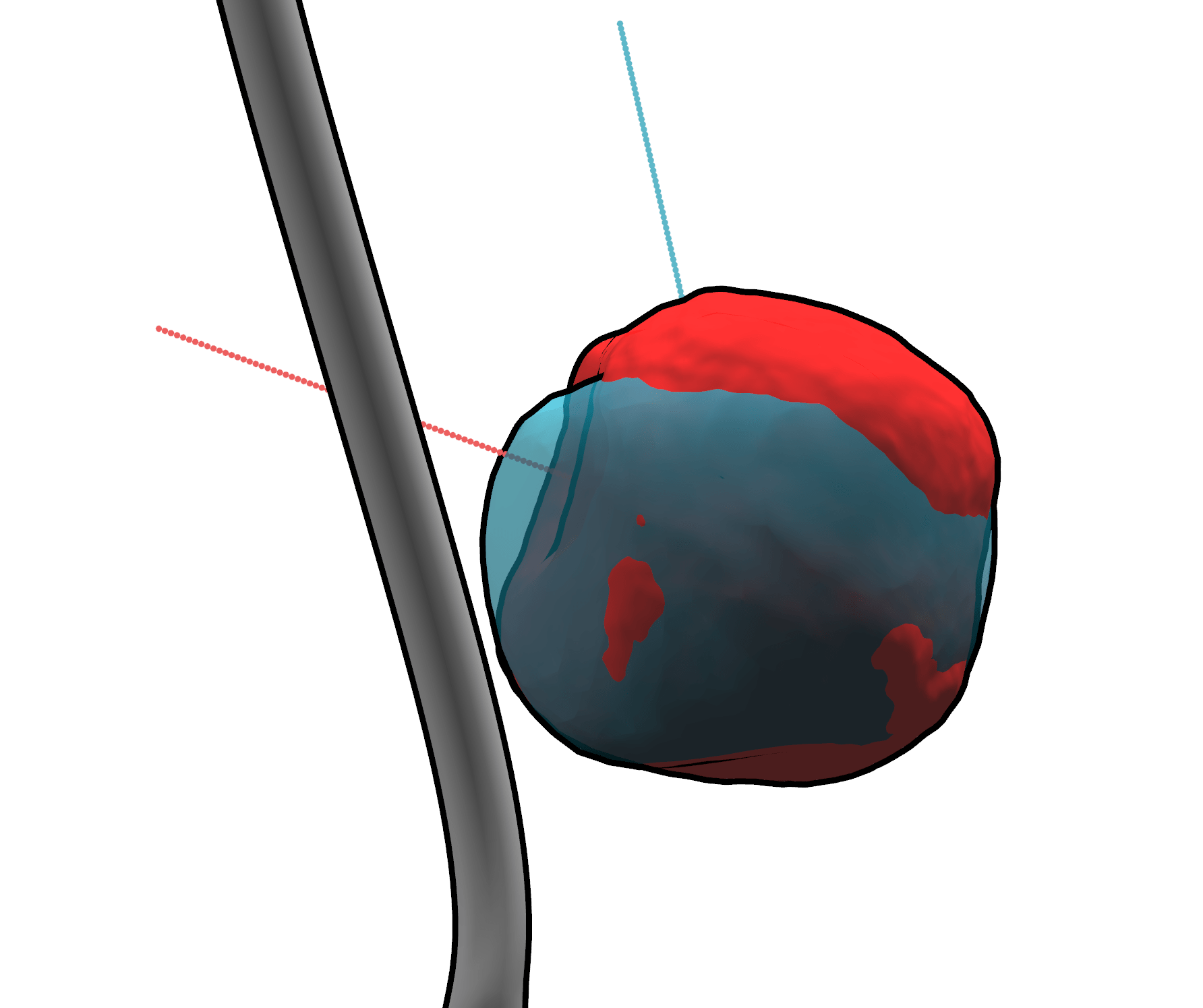} &
    \includegraphics[clip, trim=2.0cm 0cm 2.0cm 0cm, width=0.115\textwidth]{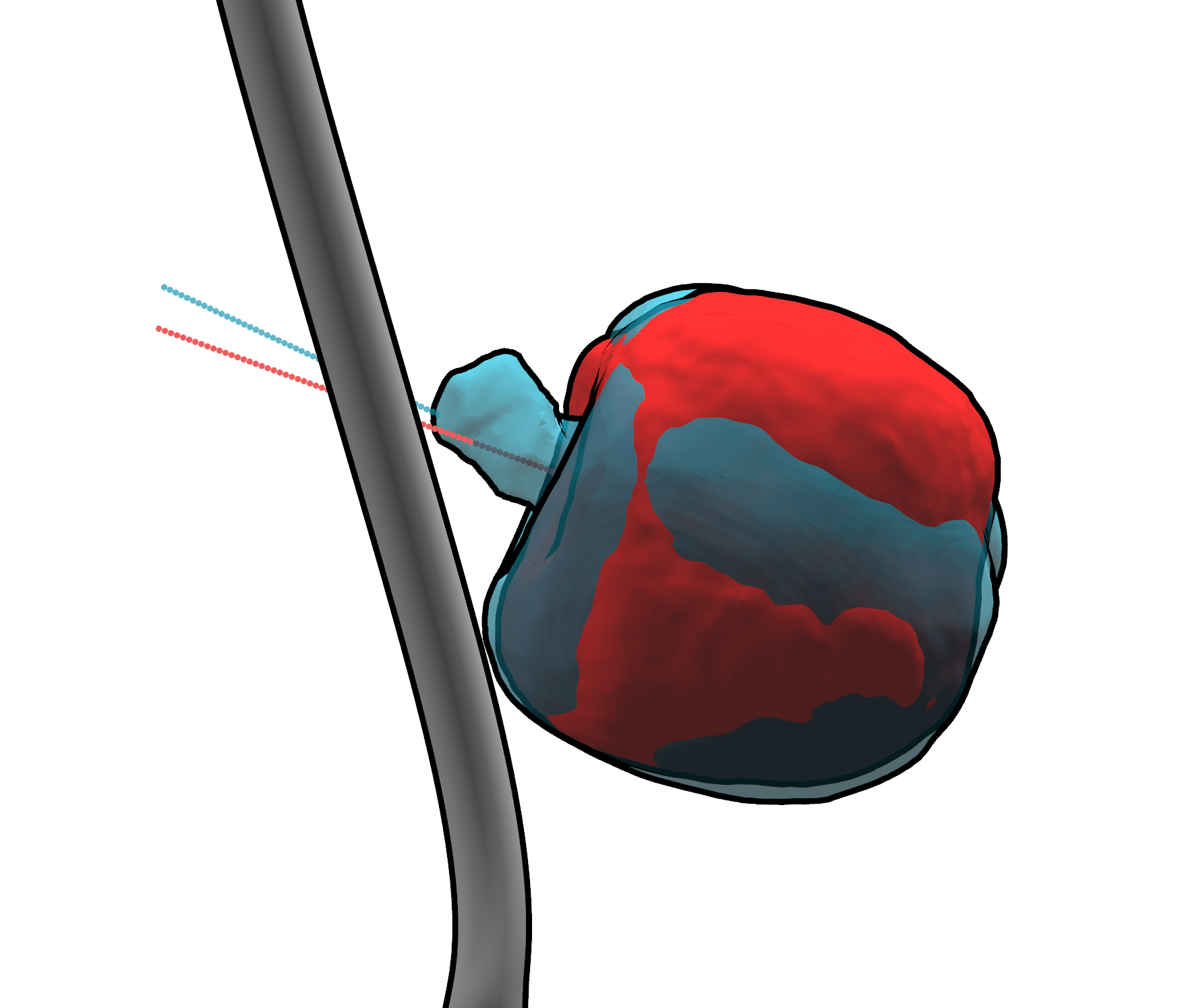} \\
    & (a) SupEps~\cite{marangoz2022case} & (b) TCoRe~\cite{magistri2024icra} & (c) HoMa~\cite{pan2023panoptic} & (d) SSC (Ours) 
    \end{tabular}    
    \caption{Illustration of shape completion with initial and without leaf occlusion. 
    Red mesh is the ground truth fruit and blue mesh indicates the estimated fruit with different methods. Red/blue lines indicate ground truth/estimated peduncle axes.
    }
    \label{fig:ShapeCompletionComparison}
    \vspace{-0.3cm}
\end{figure}

\begin{table}[!t]
    \centering
    \setlength{\tabcolsep}{2.2pt}
    \begin{tabular}{p{1.2cm}|c|cccc}
         & Method & \makecell[c]{Chamfer \\ Dis. (cm)} & \makecell[c]{Volume \\ Err. (cm$^3$)} & \makecell[c]{Center \\ Err. (cm)} &  \makecell[c]{Angle \\ Err. (deg)} \\
         \hline
         \multirow{4}{*}{\shortstack{Initial \\ Occlusion}} 
         & SupEps~\cite{marangoz2022case} & 2.68 & 224.92 & 2.40 & 36.50 \\
         & TCoRe~\cite{magistri2024icra} & 2.60 & 77.26 & 2.59 & 51.63 \\
         & HoMa~\cite{pan2023panoptic} & 2.04 & 108.87 & 1.85 & 52.19 \\
         & SSC\,(Ours) & \textbf{1.30} & \textbf{72.62} & \textbf{1.06} & \textbf{16.64} \\
         \hline
         \multirow{4}{*}{\shortstack{Without \\ Occlusion}} 
         & SupEps~\cite{marangoz2022case} & 2.67 & 384.14 & 1.73 & 28.59 \\
         & TCoRe~\cite{magistri2024icra} & 1.23 & 57.71 & 0.97 & 51.63 \\
         & HoMa~\cite{pan2023panoptic} & 1.06 & 60.29 & 0.78 & 35.49 \\
         & SSC\,(Ours) & \textbf{0.64} & \textbf{12.18} & \textbf{0.30} & \textbf{12.00} \\
         \hline
    \end{tabular}
    \caption{Evaluation of shape completion accuracy with initial and without leaf occlusions.
    The accuracy is quantified using Chamfer Distance, Volume Error, Center point distance Error, and peduncle-axis Angle Error.
    Each value is the averaged mean across 12 test scenarios.
    As can be seen, (1) the accuracy of shape completion methods generally improves with increased visibility; (2) our SSC method outperforms all baselines across all metrics.
    }
    \label{tab:ShapeCompletionStats}
    \vspace{-0.7cm}
\end{table}

\begin{figure}[!t]
    \centering
    \scriptsize
    \setlength{\tabcolsep}{0.8pt}
    \begin{tabular}{ccc}
    \includegraphics[width=0.12\textwidth]{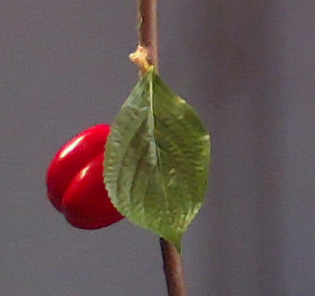} & 
    \includegraphics[width=0.12\textwidth]{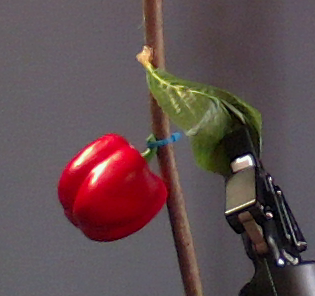} & 
    \includegraphics[width=0.12\textwidth, trim=0cm 0cm 5cm 0cm, clip]{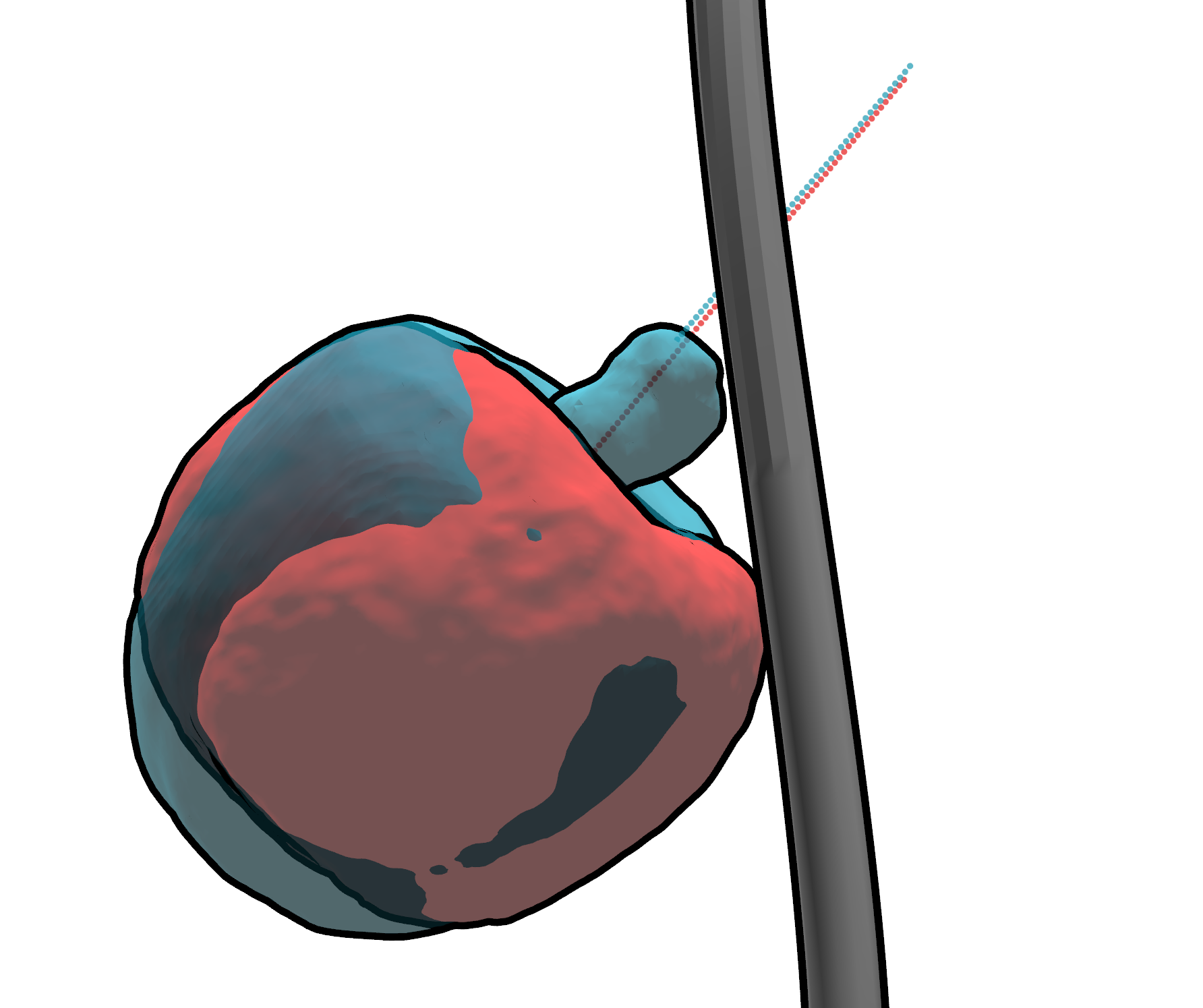} 
    \\
    (a) Initial & (b) LMAP (Ours) & (c) Final Estimation  \\ 
    \includegraphics[width=0.12\textwidth]{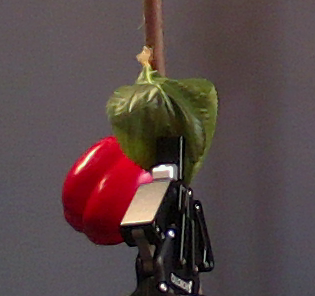} &
    \includegraphics[width=0.12\textwidth]{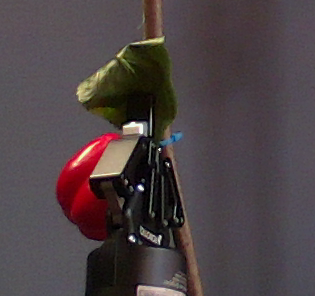} &
    \includegraphics[width=0.12\textwidth]{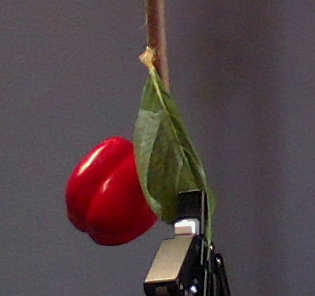}
    \\
    (d) Random & (e) Upward & (f) Min-Overlap \\ 
    \end{tabular} 
    \caption{Illustration of (a) the initial scenario, (b, d-f) different leaf manipulation actions, and (c) the final estimation of our method.}
    \label{fig:CompareDifferentActions}
    \vspace{-0.3cm}
\end{figure}

\begin{table}[!t]
    \centering
    \begin{tabular}{p{0.8cm}|c|ccc|c}
         & \makecell[c]{Move \\ Distance} & \makecell[c]{Safe \\ Rate} & \makecell[c]{Plant \\ Hazard} 
         & \makecell[c]{Fruit \\ Collision} &\makecell[c]{Visibility \\ Fraction } \\
         \hline
         \multirow{4}{*}{\shortstack{Random}}
         & 2\,cm & 46 / 60 &  1 / 60 & 13 / 60 & 0.56 $\pm$ 0.29 \\
         & 4\,cm & 28 / 60 & 18 / 60 & 14 / 60 & 0.54 $\pm$ 0.29 \\
         & 6\,cm & 22 / 60 & 23 / 60 & 15 / 60 & 0.53 $\pm$ 0.28 \\
         & 8\,cm & 18 / 60 & 25 / 60 & 17 / 60 & 0.54 $\pm$ 0.29 \\
         \hline
         \multirow{4}{*}{\shortstack{Upward}}
         & 2\,cm & 11 / 12 & 0 / 12 & 1 / 12 & 0.54 $\pm$ 0.22 \\
         & 4\,cm & 11 / 12 & 0 / 12 & 1 / 12 & 0.59 $\pm$ 0.23 \\
         & 6\,cm & 11 / 12 & 0 / 12 & 1 / 12 & 0.65 $\pm$ 0.25 \\
         & 8\,cm & 11 / 12 & 0 / 12 & 1 / 12 & 0.68 $\pm$ 0.27 \\
         \hline
         \multirow{4}{*}{\shortstack{Min- \\ Overlap}}
         & 2\,cm & 11 / 12 & 0 / 12 & 1 / 12 & 0.58 $\pm$ 0.28 \\
         & 4\,cm & 11 / 12 & 0 / 12 & 1 / 12 & 0.65 $\pm$ 0.35 \\
         & 6\,cm & 7 / 12 & 4 / 12 & 1 / 12 & 0.57 $\pm$ 0.29 \\
         & 8\,cm & 5 / 12 & 6 / 12 & 1 / 12 & 0.56 $\pm$ 0.30 \\
         \hline
         \makecell[c]{LMAP \\ (Ours)}
         & 6.7\,cm & \textbf{12} / 12 & \textbf{0} / 12 & \textbf{0} / 12 & \textbf{1.00} $\pm$ 0.00 \\
         \hline
    \end{tabular}
    \caption{
    Evaluation of varied action planning methods on artificial scenarios.
    The Random method is executed 5 times for each scenario.
    Visibility is normalized to the maximum, and mean and std. dev. are reported.
    Initial visibility is 0.44 $\pm$ 0.23. 
    For unsafe trials, the visibility fraction is set to the initial value.  
    Our method consistently plans safe manipulations that fully reveal the fruit.
    }
    \label{tab:ActionSelectionComparison}
    \vspace{-0.5cm}
\end{table}

\subsection{Impact of Visibility on Shape Completion Accuracy}
\label{sec:EvalShapeCompletion}

\textbf{Baselines.} We consider one template-based shape completion method using superellipsoids, denoted as SupEps~\cite{marangoz2022case}, and two learning-based methods using transformers, denoted as TCoRe~\cite{magistri2024icra} and DeepSDF, referred to as HoMa~\cite{pan2023panoptic}.

\textbf{Setup and Results.} To thoroughly evaluate shape completion performance, we set up two different cases for each scenario: one with initial leaf occlusions and one without.
The case without leaf occlusions is designed by manually moving the leaf aside to ensure maximum visibility from the camera viewpoint.
We compare our scene-consistent shape completion (SSC) method with baselines under two cases as shown in Tab.~\ref{tab:ShapeCompletionStats} and Fig.~\ref{fig:ShapeCompletionComparison}.
The results for the case without leaf occlusion are significantly better than those with initial leaf occlusions.
This underscores the importance of active occlusion removal for achieving more accurate fruit shape and pose estimation.
Furthermore, our SSC method provides a more accurate estimation against baselines, especially for the case with initial leaf occlusions.
This underscores the importance of our scene-consistent priors under heavy occlusion, which benefits subsequent safe leaf manipulation action planning by providing more reliable fruit shape and pose for improved collision detection and visibility checks.

\subsection{Evaluation of Leaf Manipulation Action Planning}
\label{sec:EvalActionPlanning}

\textbf{Baselines.} We consider one random and two heuristic baselines.
The Random baseline selects actions uniformly at random from all possible candidates.
Two heuristic methods first filter out actions that collide with visible plant points and then select the grasp point farthest from the branch.
For the moving direction, the Upward heuristic moves the end effector upward, while the Min-Overlap heuristic selects the direction that minimizes the overlap between the projected robot mesh and the leaf segmentation mask in the image.

\textbf{Evaluation Criterion.} We assess both the safety and visibility improvements of the planned actions.
For safety assessment, the hazardous situations and the fruit collisions are counted.
A hazardous action often overstretches the artificial leaf and branch and can even move the plant base (for more details, refer to our supplementary materials).
The system is immediately stopped in these situations.
Fruit collisions are also important during monitoring because we want to avoid damaging the fruit, which could reduce production.

\textbf{Action Planning Results.} To have a thorough evaluation, we compare our leaf manipulation action planning~(LMAP) method to the baselines under different move distances as shown in Tab.~\ref{tab:ActionSelectionComparison} and Fig.~\ref{fig:CompareDifferentActions}.
Our LMAP method achieves consistently safe actions and full visibility.
In contrast, the safety-focused Upward method fails to consistently provide adequate visibility, while the visibility-focused Min-Overlap method performs more unsafe actions.
Furthermore, our LMAP method dynamically adjusts the move distance, whereas greater distances increase the risk of plant damage.

\begin{table}[!t]
    \centering
    \begin{tabular}{c|cccc}
         & \makecell[c]{Chamfer \\ Dis. (cm)} & \makecell[c]{Volume \\ Err. (cm$^3$)} & \makecell[c]{Center \\ Err. (cm)} &  \makecell[c]{Angle \\ Err. (deg)} \\
         \hline
         LMAP with SSC & 0.68 & 24.44 & 0.33 & 11.28 \\
         \hline
    \end{tabular}
    \caption{Evaluation of final SSC shape completion accuracy of our LMAP method.
    As can be seen, the final accuracy is close to the situation without occlusion of the SSC method in Tab.~\ref{tab:ShapeCompletionStats}.
    }
    \label{tab:OursShapeCompletion}
    \vspace{-0.3cm}
\end{table}

\begin{table}[!t]
    \centering
    \begin{tabular}{c|ccc}
    \hline
         \makecell[c]{ Shape Completion} & \makecell[c]{Semantic \\ Segmentation} & \makecell[c]{Branch \\ Completion} & \makecell[c]{Fruit \\ Completion} \\
    \hline
         Runtime (s) & 5.56 & 3.03 & 13.76 \\
    \hline
         \makecell[c]{Action Planning} & \makecell[c]{Generation \\ \& Collision} & \makecell[c]{Deformation \\ Simulation} & \makecell[c]{Visibility \\ Checking} \\
    \hline
         Runtime (s) & 3.35 & 62.09 & 13.02 \\
    \hline
    \end{tabular}
    \caption{Runtime of each component of our LMAP method.}
    \label{tab:PlanningTime}
    \vspace{-0.5cm}
\end{table}

\textbf{Final Shape Completion.} As shown in Tab.~\ref{tab:OursShapeCompletion}, improved visibility by our robot actions does lead to more accurate estimation. 
The small discrepancy compared to the without occlusion (Tab.~\ref{tab:ShapeCompletionStats}) might be due to perception noise caused by slight plant movement following action execution.

\textbf{Planning Time.} We visualize the runtime of different components of our LMAP method in Tab.~\ref{tab:PlanningTime}.
The Random and Upward baselines take around 0.5 seconds while the Min-Overlap baseline takes around 10 seconds for overlap checking.
Our method takes around 100 seconds longer than baselines, but this additional time is justified as it improves visibility while minimizing the risk of plant damage.

\subsection{Evaluation on Real Leaves and Sweet Peppers}
\label{sec:EvalRealPlant}

\begin{figure}[!t]
    \centering
    \scriptsize
    \setlength{\tabcolsep}{0.8pt}
    \begin{tabular}{cccc}
        \includegraphics[width=0.12\textwidth]{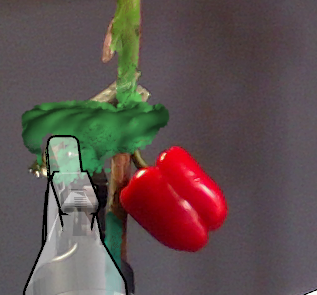} & 
        \includegraphics[width=0.12\textwidth]{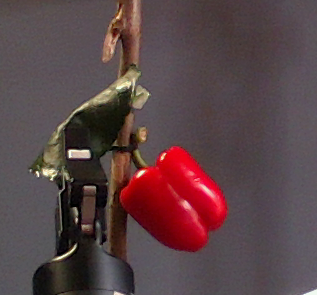} &
        \includegraphics[width=0.12\textwidth]{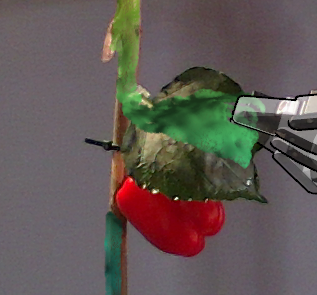} &
        \includegraphics[width=0.12\textwidth]{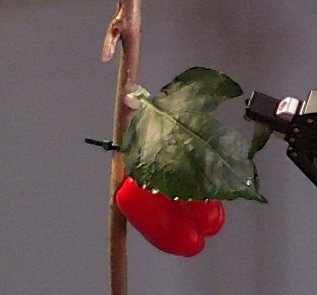} \\
        \multicolumn{2}{c}{(a) With energy} & \multicolumn{2}{c}{(b) Without energy} \\
    \end{tabular}
    \caption{
    Illustration of our LMAP method with and without energy weight. 
    In (a) and (b), the left image depicts simulated leaf deformation (green mask) and the robot surface (white mask), and the right image shows the execution result. 
    Both actions achieve good visibility in simulation; however, LMAP without energy weight selects an action that results in leaf damage during execution.
    }
    \label{fig:SuccessCase}
    \vspace{-0.3cm}
\end{figure}

\textbf{Safety Study on Real Leaves.} As real leaves are prone to damage, we further evaluate our LMAP methods on real configurations as shown in Fig.~\ref{fig:TestCases}(c).
As shown in Tab.~\ref{tab:RealLeafStudy} and Fig.~\ref{fig:SuccessCase}, planning actions solely based on maximizing visibility is insufficient to ensure safety with real leaves, underscoring the importance of considering energy.

\begin{figure}[!t]
    \centering
    \scriptsize
    \begin{tabular}{cc}
        \includegraphics[width=0.12\textwidth]{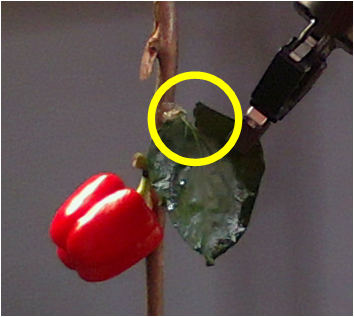} 
        &  
        \includegraphics[width=0.12\textwidth]{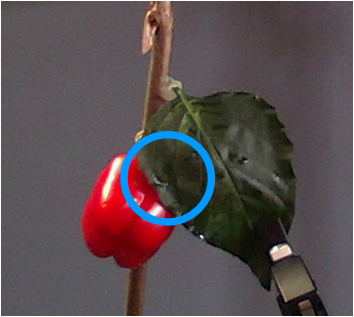} 
        \\
        (a) Leaf damage & (b) Insufficient visibility \\
    \end{tabular}
    \caption{Illustration of the two failures of LMAP reported in Tab.~\ref{tab:RealLeafStudy}.
    The yellow circle represents the area of leaf damage, while the blue area shows the fruit surfaces that remain occluded.
    }
    \label{fig:FailureCases}
     \vspace{-0.3cm}
\end{figure}

\textbf{Analysis of Failure Cases.} We observe two failure cases of our LMAP method as shown in Fig.~\ref{fig:FailureCases}. 
Leaf damage may result from slight inaccuracies in the deformation simulation, as real leaves may have more complex material properties than artificial ones.
Insufficient visibility may result from slight inaccuracies in shape completion, as real peppers may have more complex surfaces compared to artificial ones.

\begin{table}[!t]
    \centering
    \begin{tabular}{c|c|cc}
         & Success & Leaf Damage & Insufficient Visibility  \\
         \hline
         Without energy & 8/16 & 8 & 0 \\
         With energy & 14/16 & 1 & 1 \\
         \hline
    \end{tabular}
    \caption{Evaluation of LMAP method with ($\lambda=1$) and without energy weight ($\lambda=0$) on real leaves and peppers.
    Success indicates that the leaf remains undamaged and the fruit is maximally visible, annotated by a human.
    Right two columns break down failure cases.
    }
    \label{tab:RealLeafStudy}
    \vspace{-0.5cm}
\end{table}

\textbf{Multi-Fruit Scenario.} We test a real multi-fruit setup as shown in Fig.~\ref{fig:StarFigure} and video.
Our system achieves maximum visibility for all fruits while preserving integrity of all leaves.

\section{Conclusion and Future Work} \label{S:conclusions}

In this work, we present an active fruit shape and pose estimation system for autonomous crop monitoring under heavy leaf occlusion. 
To the best of our knowledge, we are the first to employ physical manipulation to move occluding leaves, allowing direct observation of hidden fruit.
Comprehensive experiments with artificial and real plants demonstrate that (1) our novel scene-consistent shape completion significantly improves estimation accuracy, and (2) our leaf manipulation action planning consistently selects robot actions that maximize visibility while minimizing leaf damage.
The multi-fruit scenario further highlights the applicability of our system.

We plan to deploy our method in a glasshouse environment using a stronger segmentation model capable of segmenting green stems and sweet peppers, integrated into the Hortibot system~\citep{lenz2024hortibot}.
Multiple arms can enable viewpoint planning for the detection of sweet peppers completely hidden by leaves and pushing aside irrelevant plant parts, thereby reducing cases where multiple leaves occlude sweet peppers.

\clearpage
\bibliographystyle{IEEEtranSN}
\footnotesize
\bibliography{ICRA2025}

\end{document}